\documentclass{article}

\usepackage{arxiv}

\usepackage{hyperref}
\usepackage{bm}
\usepackage{cite}
\usepackage{graphics,fancyhdr,graphicx,subfigure}
\usepackage{subfigure}
\usepackage{amsthm,amsmath,amsfonts,latexsym,amssymb}
\usepackage{lineno}
\usepackage{diagbox}
\usepackage[ruled,linesnumbered]{algorithm2e}
\SetKwComment{Comment}{$\triangleright$\ }{}
\usepackage{listings}
\usepackage[table]{xcolor}
\usepackage{lscape}
\usepackage{comment}
\usepackage{tabularx}
\usepackage[ruled,linesnumbered]{algorithm2e}
\usepackage{graphicx}
\graphicspath{{images/}}
\usepackage{tabularx}
\usepackage{nomencl}
\usepackage{array}
\usepackage{caption}
\usepackage{breqn}
\usepackage{lineno}
\usepackage{mathtools}
\usepackage{subfigure}
\usepackage{caption}
\usepackage{lmodern}
\usepackage{calligra}
\usepackage{xspace}
\usepackage[utf8]{inputenc}
\usepackage{enumerate}
\usepackage[english]{babel}

\def\equationautorefname~#1\null{%
  Eq.~(#1)\null
  }
\def\subfigureautorefname~#1\null{%
  Fig.~#1\null
}

\definecolor{listinggray}{gray}{0.9}
\definecolor{lbcolor}{rgb}{0.9,0.9,0.9}
\definecolor{Darkgreen}{RGB}{0,100,0}

\title{Surrogate assisted active subspace and active subspace assisted surrogate - A new paradigm for high dimensional structural reliability analysis\thanks{https://www.csccm.in/}}

\author{ \hspace{1mm}Navaneeth~N.\\
	Department of Applied Mechanics\\
	Indian Institute of Technology (IIT) Delhi\\
	Hauz Khas - 110 016, New Delhi, India \\
	\texttt{navaneeth.n@am.iitd.ac.in} \\
	\And
	\hspace{1mm}Souvik~Chakraborty \\
	Department of Applied Mechanics\\
	Indian Institute of Technology (IIT) Delhi\\
	Hauz Khas - 110 016, New Delhi, India \\
	\texttt{souvik@am.iitd.ac.in} \\
}

\renewcommand{\shorttitle}{Sparse active subspace based H-PCFE}

\hypersetup{
pdftitle={SAS based H-PCFE},
pdfsubject={q-bio.NC, q-bio.QM},
pdfauthor={Navaneeth N., Souvik Chakraborty},
pdfkeywords={Active subspace, Surrogate Model, Manifold learning},
}

\begin{document}
\maketitle

\begin{abstract}
	Performing reliability analysis on complex systems is often computationally expensive. In particular, when dealing with systems having high input dimensionality, reliability estimation becomes a daunting task. A popular approach to overcome the problem associated with time-consuming and expensive evaluations is building a surrogate model. However, these computationally efficient models often suffer from the \textit{curse of dimensionality}. Hence, training a surrogate model for high-dimensional problems is not straightforward. Henceforth, this paper presents a framework for solving high-dimensional reliability analysis problems. The basic premise is to train the surrogate model on a \textit{low-dimensional manifold}, discovered using the active subspace algorithm. However, learning the low-dimensional manifold using active subspace is non-trivial as it requires information on the gradient of the response variable. To address this issue, we propose using sparse learning algorithms in conjunction with the active subspace algorithm; the resulting algorithm is referred to as the sparse active subspace (SAS) algorithm. We project the high-dimensional inputs onto the identified low-dimensional manifold identified using SAS. A high-fidelity surrogate model is used to map the inputs on the low-dimensional manifolds to the output response. We illustrate the efficacy of the proposed framework by using three benchmark reliability analysis problems from the literature. The results obtained indicate the accuracy and efficiency of the proposed approach compared to already established reliability analysis methods in the literature.
\end{abstract}


\section{Introduction}
\label{sec:intro}
It is well known that uncertainties inherently exist in practical engineering systems in various forms, such as randomness associated with the geometrical configurations, material properties, boundary conditions and loading conditions. Therefore, quantitative assessment of reliability becomes extremely important in practice \cite{ditlevsen1996structural}. Though the evaluation of failure probability can be analytically formulated as a multi variate integral \cite{haldar2000reliability,haldar2000probability} in the failure domain, computation of the integral often becomes infeasible due to irregularities in the domain and high dimensionality of the inputs. A possible alternative is to use numerical integration schemes(e.g., Monte Carlo integration); however, the computational cost becomes huge for practical engineering problems.
Therefore, one of the main research focuses among the structural reliability analysis research community is development of efficient tools that can solve high-dimensional reliability analysis problems.
\par
The most popular method for reliability analysis is perhaps the Monte Carlo simulation (MCS) \cite{thakur1978monte,rubinstein2016simulation}.
MCS is a widely used method for computing multivariate integral in statistical physics. It performs simulations for the large number of sample points drawn independently from the probability distribution of the input variables. Therefore, in principle, the method can be employed to estimate failure probability by enumerating simulation results. Although the procedure is simple and straight forward, MCS becomes quite expensive as it has to utilize a large number of simulations to assure the convergence of a solution. To overcome the aforementioned problem, researchers have developed methods that enhance the computational efficiency of the crude MCS. For example, importance sampling \cite{au1999new,li2005curse,engelund1993benchmark}, subset simulations \cite{au2001estimation,au2014engineering,zuev2015subset} and directional simulations \cite{ditlevsen1990general} have superior convergence rate as compared to crude MCS; however, these techniques still require a considerable number of simulations to achieve accurate estimations.
\par 
Analytical approximation-based approaches, an alternative to sampling-based approaches discussed above, are often chosen to evaluate reliability. Here, the approximation of the multivariate integral of limit state function over the failure domain is evaluated utilizing the Taylor's series expansion and asymptotic methods. 
The most popularly analytical approximation based methods are perhaps the
First-Order Reliability Method (FORM) \cite{hohenbichler1987new,zhao1999general,hu2015first} and Second-Order Reliability Method (SORM) \cite{zhang2010second,lee2012novel}.
While FORM uses first-order Taylor's series expansion, SORM utilizes a second-order Taylor's series expansion.
Improvements to conventional FORM and SORM have also been pursued by the researchers \cite{kiureghian1991efficient,koyluoglu1994new}.
These analytical approximation based methods are generally computationally efficient; however, the results are often not accurate for highly nonlinear systems.
\par 
Surrogate based approaches have recently gained wide popularity in analyzing reliability of structures due to their comparable accuracy and cost effective computation. In this method, the actual limit state function is replaced by a machine learning model. The training data necessary for training the machine learning model is generated by using a design of experiment scheme \cite{chakraborty2016sequential,bhattacharyya2018critical}. Polynomial
chaos expansion \cite{blatman2011adaptive,sudret2008global}, radial basis function (RBF)  \cite{de2013new,li2018sequential},  Gaussian process \cite{bilionis2012multi,bilionis2013multi}, neural networks \cite{chakraborty2020simulation,chakraborty2021transfer}
and support vector machines \cite{dai2012structural,guo2009application,roy2020support,ghosh2018support} are some of the widely used surrogate models. Hybrid surrogate models \cite{schobi2015polynomial,chakraborty2017efficient,chakraborty2017hybrid} are also often used in reliability, where the models are developed by incorporating more than one surrogate models as to eliminate inherent limitations of different components.
Though a surrogate model works well for low-dimensional systems, constructing an accurate model becomes challenging with an increase in the dimensionality of input variables.
\par
As was pointed out in the above discussion, dealing with high dimensional inputs is one of the persisting problems of surrogate modelling. The traditional way to tackle this problem is to use dimensionality reduction techniques to identify a low-dimensional manifold; subsequently, the surrogate model is trained in the low-dimensional space. 
Dimensionality reduction methods available in the literature are commonly classified into two main categories: feature selection and feature extraction \cite{hira2015review}. 
While the feature selection methods seek the set of most influential input variables \cite{saltelli2008global}, feature extraction methods transform the original higher-dimensional space into a subspace of reduced dimension. The main limitation of the feature selection method is that it fails when all input variables have more or less equal importance to the quantity of interest. However, feature extraction methods circumvent this limitation as they utilize a combination of all input variables. Principle component analysis is a widely used feature extraction method \cite{jolliffe2016principal}, where the subset is approximated based on the information contained merely in the input variables space. Since these methods do not account for information associated with the target data (quantity of interest), they may not effectively reduce the dimension. Consequently, supervised reduction techniques have received much attention from the scientific community. One such supervised dimensionality reduction technique is the active subspace method \cite{constantine2014active,constantine2014computing}. This method discovers the subspace through gradient evaluations of the output variable. Usually, finite difference or related techniques are adopted to estimate the gradient of a given function. 
Unfortunately, direct application of this technique for solving reliability analysis problem is challenging as the number of function evaluations increases significantly with an increase in the input dimensionality.
\par
Given the drawbacks mentioned above associated with reliability analysis and dimensionality reduction techniques, 
we propose a novel reliability analysis framework that couples surrogate models with the active subspace algorithm wherein (a) the surrogate assists active subspace in efficiently computing the gradients and the (b) active subspace assists in efficiently training the surrogate model.
The proposed framework has three components 
(a) a sparse and low-fidelity surrogate model,
(b) the active subspace algorithm, and 
(c) a high-fidelity surrogate model.
The sparse and low-fidelity surrogate model assists the active subspace algorithm in efficiently computing the gradient information. We refer to this combination of sparse surrogate model and active subspace as the sparse active subspace (SAS).
Finally, the high-fidelity surrogate model is used to map the SAS discovered inputs in the low-dimensional manifold and the output.
While any sparse surrogate model can be used as a low-fidelity surrogate, we propose to use the least angle regression-based sparse polynomial chaos expansion proposed in \cite{blatman2011adaptive}.
On the other hand, we propose to use Hybrid Polynomial Correlated Function Expansion (H-PCFE) \cite{chakraborty2017efficient,chakraborty2017moment,chatterjee2016bi} as the high-fidelity surrogate.
We illustrate that the proposed approach is highly efficient in solving high-dimensional reliability analysis problems. As for accuracy, the proposed framework outperforms popular reliability analysis techniques such as FORM and SORM. We also illustrate that results obtained using the framework proposed is significantly more accurate as compared to direct application of sparse surrogate model for reliability analysis.
\par
The rest of the paper is organized as follows. In Section \ref{sec:pbs}, the general problem setup is presented. Section \ref{sec:pa} elucidates the proposed approach. This is followed by three numerical examples, illustrated utilizing the proposed approach, in Section \ref{sec:ne}. Finally, the concluding remarks are provided in Section \ref{sec:conclusions}.

\section{Problem statement}\label{sec:pbs}
Consider an N-dimensional vector of random variables $\bm{X}$, $\bm{X}=\left(X_{1},X_{2},....,X_{N}\right)$ : $\Omega_{\bm X}\to \mathbb{R}^{N}$, with probability density function $P_{\bm X}(\bm x)$ and cumulative distribution function $F_{\bm{X}}\left(\bm{x}\right)=\mathbb P \left(\bm{X}\leq \bm{x}\right)$, where $\mathbb P$ denotes the probability and $\Omega_{\bm X}$ denotes the probability space. Failure probability of a given system is quantified based on the limit state function $\jmath(\bm x) = 0$. The function describes the reliability of the system such that $\jmath(\bm x)<0$ denotes the failure domain $\Omega_{\bm X}^{F}$ 
\begin{equation}
\Omega_{\bm X}^{F} \overset{\Delta}{=}\{\bm x :\jmath(\bm x)<0\},
\end{equation}
and $\jmath(\bm x) > 0 $ denotes the safe region. $\jmath(\bm x) = 0 $ represents the limiting condition.
The failure probability is defined as
\begin{equation}
P_{f}=\mathbb P(\bm{X}\in \Omega_{\bm{X}}^{F})=\int_{\Omega_{\bm{X}}^{F}}dF_{\bm{X}}(\bm x)=\int_{\Omega_{\bm{X}}}{\xi_{\Omega_{\bm{X}}^{F}}dF_{\bm{X}}(\bm{x})},
\end{equation}
where $\xi_{\zeta}$ is a characteristic function and satisfies.
 
\begin{equation}
    \xi_{\zeta}(\bm{x})=
    \begin{cases}
      0, & \text{if}\ \bm{x}\in \zeta \\
      1, & \text{otherwise}
    \end{cases}.
  \end{equation}
The vitality of the limit function $\jmath(\bm{x})$ in determining the probability of failure is clearly vindicated.
Here, the generalised multivariate form of limit state function can be represented as
$y=\jmath(\bm{x}),\jmath:\mathbb{R}^{N}\to\mathbb{R}$, with $N \ge 1$. Evaluation of the reliability is a computationally intensive process, as it involves a great number of the function evaluations. However, an approximation of $\jmath(\bm{x})$ can be constructed using a surrogate model, $\hat{\jmath}(\bm{x})$ which learns an approximate mapping between the inputs and the output by using training samples generated using design of experiments $\bm{\Xi}=\left[\bm X^{(1)},\bm X^{(2)},\ldots,\bm X^{(N_s)}\right]^T$ to corresponding function evaluations $\bm Y = \left[Y^{(1)}, Y^{(2)}, \ldots, Y^{(N_s)}\right]^T$, where $N_s$ represents the number of training samples. Mathematically,
\begin{equation}
\jmath(\bm{x})=\hat{\jmath}(\bm{x};\bm{\theta})+\epsilon,
\end{equation}
where $\bm{\theta}$ denotes the set of surrogate parameters and $\epsilon$ denotes the surrogate error.
Often the construction of $\hat{\jmath}$ with actual dimension may not be computationally feasible. Thus a reduced dimension is used to construct the surrogate model. The reduced subspace can be represented as $\bm z = h_{Nr}:\mathbb{R}^{N}\to\mathbb{R}^{N_r}$,and the surrogate can be represented as
\begin{equation}
\jmath(\bm x)=\tilde{\jmath}(\bm{z};\theta)+\epsilon.
\end{equation}
One challenge associated with reliability analysis is to reduce the number of actual simulations, $N_s$ as number of simulations is directly proportional to the computational cost.
Similarly, the goal of reduced order model is to reduce the effective dimensionality $N_r$ as reduces the computational cost associated with training a surrogate model.
The objective of this paper is to propose a novel reliability analysis framework that reduces the number of effective dimensions $N_r$ as well as the number of training samples $N_s$.
Details on the proposed framework along with its different components are furnished in the next section.

\section{Proposed approach}\label{sec:pa}
In this section, we provide details about the reliability analysis framework proposed in this paper. However, before going into the details of the proposed approach, we briefly review active subspace, sparse polynomial chaos expansion and H-PCFE. These three are the key ingredients of the proposed approach.
\subsection{Active subspace}
Active subspace is a novel reduced-order-modelling method that effectively approximates higher dimensional physical models to a lower-dimensional model. The method transforms the original coordinate so that the input parameters are aligned to the direction of strongest variability.
\par
We assume ${f}$ is continuous and differentiable over $\bm{x}$ with a probability density function $\rho$. The gradient vector of ${f}$ is represented as
${\nabla_{\bm{x}}}f(\bm{x})=\left[\frac{\partial f}{\partial x_{1}},\frac{\partial f}{\partial x_{2}},....,\frac{\partial f}{\partial x_{N}}\right]^{T}$ and is used to evaluate average derivative functional $\mathbf C$ \cite{constantine2014computing,constantine2014active}
\begin{equation}\label{C}
\textbf{C}=\int({\nabla_{\bm{x}}}f)({\nabla_{\bm{x}}}f)^T{\rho}d{\bm{x}}.
\end{equation}
$\mathbf C$ is a positive semi-definite matrix. Using eigen value decomposition, we have
\begin{equation}\label{decomposition}
\mathbf{C}=\mathbf{W}{\mathbf{\Lambda}}\mathbf{W}^T, 
\end{equation}
where 
\begin{equation}
    \mathbf{\Lambda}={diag}(\lambda_{1}, \lambda_{2},\ldots,\lambda_{N}),
\end{equation}
such that  $\lambda_{1}\geq\lambda_{2}\geq\cdots\geq \lambda_{N}\geq{0}$.
Further, the eigen vector space $\mathbf{W}$ can be splitted into two parts $\mathbf{W_{1}}, \mathbf{W_{2}}$, and corresponding eigen values to $\mathbf{\Lambda_{1}} ,\mathbf{\Lambda_{2}}$ 
\begin{equation}\label{partition}
\mathbf{W}=[\mathbf{W_{1}},\mathbf{W_{2}}], \quad \mathbf{\Lambda}=\begin{bmatrix}
\mathbf{\Lambda_{1}} & \\
 &\mathbf{\Lambda_{2}}
\end{bmatrix}.
\end{equation}
$\mathbf{\Lambda_{1}}$ contains the first $r$ largest eigen values, corresponding eigen vector subspace $\mathbf{W_{1}}\in \mathbb{R}^{r}$ is called active subspace, which contains the first $r$ orthonormal eigen vectors such that
\begin{align}\label{Threshold}
\frac{\sum_{i=1}^{r} {\lambda_{i}}}{\sum_{i=1}^{n}{\lambda_{i}}}\leq \mu,
\end{align} 
where $\mu$ is the threshold value and is used to partition the active subspace.
Once the active space is defined function $f$ can be approximated as
\begin{equation}
{f}\approx\hat{f}=f(\mathbf{{W}}_1^{T}\bm{X})=f(\bm{Z}),
\end{equation} 
where $\bm Z=f(\mathbf{{W}}_1^{T}\bm{X})$.
In practice, to approximate the higher dimensional integration in the construction of $\mathbf C$, Monte Carlo integration is employed. The approximate of $\mathbf C$ with $N_{s_1}$ number of samples is given by
\begin{equation}\label{montecarlo}
    \hat{\mathbf{C}}\approx\frac{1}{N_{s_1}}\sum_{j=1}^{N_{s_1}}({\nabla_{\bm{x}}}{f}_{j})({\nabla_{\bm{x}}}{f_{j}})^T.
\end{equation}
A step-by-step procedure of computing active subspace is shown in Algorithm \ref{alg:ASS}.

\begin{algorithm}[ht!]
 \caption{Active Subspace}\label{alg:ASS}
 \textbf{Initial sampling:} Draw a set of \textit{M} samples, each $x_{j}$ independently from $\xi_{\zeta}$.\\
 Compute gradient ${\nabla_{\bm{x}}}{f}_{j}(\bm{x})$ for each $x_{j}$ .\\
 Construct the average derivative functional matrix $\hat{\mathbf{C}}$  \Comment*[r]{\autoref{montecarlo}} 
 Compute eigen value decomposition of $\hat{\mathbf{C}} \to \mathbf{W}\mathbf{\Lambda}\mathbf{W}^T $\Comment*[r]{\autoref{decomposition}} 
 Use threshold values  to partition the eigen vector space \Comment*[r]{\autoref{partition}} 
\end{algorithm}
\noindent
\textbf{Remark 1:} Although active subspace algorithm is highly effective in reducing the effective dimensionality, the computational cost associated with this algorithm is significant. Considering $N_{s_1}$
to be the number of samples, $N$ to be number of variables, the number of function evaluations $N_f$, required in active subspace is
\begin{equation}\label{eq:func_eval_ASS}
    N_f = N_{s_1} \left( N + 1 \right).
\end{equation}
It is easy to follow that $N_f$ will become huge for high-dimensional systems. This prohibits direct application of active subspace to high-dimensional reliability analysis problems with computationally expensive limit-state function.

\subsection{Sparse polynomial chaos expansion}
This section discusses the second component of the proposed framework, namely the sparse polynomial chaos expansion (S-PCE).
Among different variants of S-PCE available in the literature, we have used least angle regression (LAR) based S-PCE proposed in \cite{blatman2011adaptive}. Accordingly, we only discuss LAR based S-PCE in this section.
\par 
Consider a set of random independent input variables $\bm X \in \mathbb{R}^{N}$ and output variables $Y \in \mathbb{R}$ of a system. The mapping between the input and output variables is denoted as $\mathcal{M}: \bm X \mapsto Y$, which in generally can be a computational model (e.g., FE model). In LAR based S-PCE, the mapping $\hat{\mathcal{M}}$ is defined as \cite{blatman2011adaptive}
\begin{equation}\label{eq:10}
     y\approx\hat{\mathcal{M}}(\bm x)=\sum_{\beta\in{\mathbb{N}^{N}}}{a}_{\beta}{\psi}_{\beta}(\bm{x}),
\end{equation}
where ${\psi}_{\beta}(\bm{x})$'s are multivariate polynomials and $a_{\beta}$ are the unknown coefficients. $\bm x$ is a realization of vector $\bm X$ and $y$ is the corresponding realization of output variable $Y$. The constructed series given in the Eq.\eqref{eq:10} converges in the $L^{2}$ sense. 
Since the family of multivariate polynomials in the PCE are orthogonal, it follows the condition
\begin{equation}\label{eq:11}
     \mathbb{E}(\psi_{i}(\bm{X})\psi_{j})(\bm{X})=\langle \psi_{i}(\bm{X}),\psi_{j}(\bm{X})\rangle=\delta_{ij},
\end{equation}
where $\delta_{ij}=1$ if i=j and 0 otherwise.
In general, projection and regression are principle strategies adopted to estimate the coefficients of PCE non-intrusively. 
The number of terms in PCE grows factorially with an increase in the number of random variables, and hence,
PCE suffers from the so-called \textit{curse of dimensionality}.
To circumvent this issue, S-PCE only retains the effective/influential terms of PCE.
In this context, LAR is one of the promising variable selection methods available in the literature.
In LAR based S-PCE, one uses LAR to only retain the important terms in the PCE expression.
LAR chooses the actives set of basis functions that have strongest correlation with the model response, and it further determines the best set of coefficients iteratively using analytical relations. Description of the steps involved in the procedure of LAR based PCE is shown in Algorithm \ref{alg:LAR}.

\begin{algorithm}[ht!]
\caption{LAR based PCE}\label{alg:LAR}

\textbf{Initialize:} Initialize all unknown coefficients $a_{\beta}=0$. Set the      initial     residual values($r_{0}$) to y and active set to null set($\phi$).\\
Find the vector ${\psi_{\beta}}_{j}$ which has most  correlation with the residual.\\
Move all coefficients of active set ${a_{\beta}}_{j}$ from 0 to the least square    solution of the current residual on ${\psi_{\beta}}_{j}$ until some other predictor ${\psi_{\beta}}_{j}$  is equicorrelated to the current residual as does ${\psi_{\beta}}_{j}$.\\
Move the combined set of coefficients $\{{a_{\beta}}_{j},{a_{\beta}}_{k}\}$ towards  the joint least square solutions of $\{{\psi_{\beta}}_{j},{\psi_{\beta}}_{k}\}$ until some other predictor is equicorrelated with the current residual.\\
Continue the previous steps until $m=min(P,N-1)$ predictors have been entered  where m is the size of the active set, P is maximum order polynomial and N is the number of variables.   
\end{algorithm}
\par
\noindent
\textbf{Remark 2:} S-PCE, although highly efficient, often does not yield adequately accurate results for reliability analysis problems. This is because PCE is a global surrogate model \cite{schobi2015polynomial,goswami2019threshold} represented by using global basis functions.
Reliability predictions, on the other hand, are only dependent on the tail of the response PDF and can be considered as a local measure. 

\subsection{Hybrid polynomial correlated function expansion}
The final component of the proposed framework is a novel high-fidelity surrogate model referred to as the Hybrid Polynomial Correlated Function Expansion (H-PCFE). The method adopts a hybrid approach, where the Polynomial Correlated Function Expansion (PCFE) is combined with the Gaussian process (GP). Thus, H-PCFE performs on two levels: firstly, the global behaviour is approximated by the PCFE, and secondly, the local variations are interpolated by GP based on a covariance kernel. 
As a result, results obtained using H-PCFE is generally more accurate.
\par 
Suppose, $\bm{X}=\left(X_{1},X_{2},....,X_{N}\right)$ : $\Omega_{X}\in \mathbb{R}^{N}$ to be the inputs as before and $Y$ to be the output. H-PCFE represents the input output mapping as 
\begin{equation}\label{eq:14}
    {Y}=f_{\texttt{PCFE}}+ f_{\texttt{GP}},
\end{equation}
where $f_{\texttt{PCFE}}$ represents the functional form of PCFE \cite{chakraborty2015polynomial, chakraborty2017towards,chakraborty2017polynomial} and $f_{\texttt{GP}}$ represents a zero mean GP \cite{nayek2019gaussian} respectively.
Using extended bases, PCFE can further be expressed as
\begin{equation}\label{eq:15}
    f_{\texttt{PCFE}}\approx {\tilde{f}_{\texttt{PCFE}}}=g_{0}+\sum_{k=1}^{M}\Bigg\{\sum_{i_{1}=1}^{N-k+1}....\sum_{i_{k}=i_{k}-1}^{N}\sum_{r=1}^{k}\Bigg(\sum_{m_{1}=1}^{b}....\sum_{m_{r}=1}^{b}{\alpha}_{m_{1}....m_{r}}^{({i_{1}}{i_{2}}...{i_{k}}){i_{r}}}{{\psi}_{m_{1}}^{i_{1}}....{\psi}_{m_{r}}^{i_{r}}}\Bigg)\Bigg\},
\end{equation}
where $\psi$  represent the basis functions $\bm \alpha$ represent the unknown coefficients associated with the bases. $g_{0}$ is a constant term representing mean response.
In general, any appropriate basis functions can be chosen to form an extended bases for the PCFE. However, for faster convergence, orthogonal basis functions are used.
On the other hand,
\begin{equation}
    {f}_{\texttt{GP}}={\sigma}^{2}{Z}(\bm 0, k(\cdot,\cdot;\bm \theta),
\end{equation}
where $k(\cdot,\cdot;\bm \theta)$ represents covariance kernel function and $\bm \theta$ denotes the length scale parameter of GP. 
For using H-PCFE in practice, the unknown coefficients $\bm \alpha$, the length scale parameters $\bm \theta$ and the process variance $\sigma^2$ need to be computed. 
We use maximum likelihood estimation (MLE) to compute $\bm \alpha$, $\sigma^2$ and $\bm \theta$. This yields
\begin{equation}\label{eq:17}
     (\mathbf{\Psi}^{T}\mathbf{R}^{-1}\mathbf{\Psi})\bm{\alpha}=\mathbf{\Psi}^{T}\textbf{R}^{-1}\bm{d},
\end{equation}
and
\begin{equation}\label{eq:sigma}
    {\sigma}^{2}=\frac{1}{N_{s}}(\bm{d}-\mathbf{\Psi}\bm{\alpha})^{T}\mathbf{R}^{-1}(\bm{d}-\mathbf{\Psi}\bm{\alpha}),
\end{equation}
where $\bm \Psi$ represents the basis function matrix (a.k.a. design matrix) obtained using the training inputs and the basis functions and 
$\mathbf R$ represents the covariance matrix formulated using the training inputs and the covariance kernel,
$k(\cdot,\cdot;\bm \theta)$. $\bm d$ in Eqs. \eqref{eq:17} and \eqref{eq:sigma} is the difference between observed response of training points and mean response,  
\begin{equation}
    \bm{d}=\bm{y}-g_{0}.
\end{equation}
However, for the length-scale parameters, there exists no closed-form solution and one has to utilize numerical optimization to maximize the data likelihood.
For further details, interested readers may refer \cite{chatterjee2016bi}.

We rewrite Eq.\eqref{eq:17} in matrix form.
\begin{equation}\label{eq:18}
    \mathbf{A}\bm{\alpha}=\bm{B},
\end{equation}
where $\mathbf{A}=(\mathbf{\Psi}^{T}\mathbf{R}^{-1}\mathbf{\Psi})\bm{\alpha}$ and $\mathbf{B}=\mathbf{\Psi}^{T}\textbf{R}^{-1}\bm{d}$.
Since, we use extended bases within the PCFE, the matrix $\mathbf A$ and vector $\bm B$ have redundants. Removing the redundants, we have,
\begin{equation}\label{eq:19}
    \mathbf{A'}\bm{\alpha}=\bm{B'}.
\end{equation}
Eq. \eqref{eq:19} represents an underdetermined set of equations and has infinite solutions.
One possible alternative is to select the solution that minimizes the $L^2$-norm,
\begin{equation}\label{eq:20}
    \bm{\alpha}_{0}=\mathbf{A'}^{\dagger}\bm{B'},
\end{equation}
where $\mathbf{A'}^{\dagger}$ represents  pseudo-inverse of $\mathbf{A'}$ and satisfies all four Penrose condition.
However, this fails to satisfy the hierarchical orthogonality condition associated with PCFE \cite{chakraborty2017towards}.
To that end, a novel algorithm referred to as the homotopy algorithm (HA) is used for computing the unknown coefficients \cite{li2010d,li2012d}.
In HA, we define an additional objective function in by using a weight matrix $\mathbf W_{HA}$ that forces the solution to satisfy the hierarchical orthogonality condition.
The final solution solution obtained using HA is as follows
\begin{equation}
    \bm{\alpha_{HA}}=[\mathbf{V}_{q-r}(\mathbf{U}_{q-r}^{T}\mathbf{V}_{q-r})^{-1}\mathbf{U}_{q-r}^{T}]\bm{\alpha}_{0}.
\end{equation}
Here $\mathbf{U}_{q-r}$ and $\mathbf{V}_{q-r}$ are the last ($q-r$) rows of $\mathbf{U}$, $\mathbf{V}$ matrices, while the $\mathbf{U}$ and $\mathbf{V}$ are obtained from the singular value decomposition of $\mathbf{P}\mathbf{W}_{HA}$
\begin{equation}
      \mathbf{PW}_{HA}=\mathbf{U}\begin{bmatrix}
\mathbf{D_{r}} & 0\\
0 & 0\end{bmatrix}\mathbf{V}^{T}
  \end{equation}
\begin{equation}
    \mathbf{P}=[\mathbb{I}-(\mathbf{A}')^{-1}\mathbf{A'}]
\end{equation}
For further details on weight matrix used in H-PCFE, interested readers may refer \cite{chatterjee2016bi}.
For details on HA, interested readers may refer \cite{li2010d,li2012d}.
An algorithm depicting the steps involved in training H-PCFE is shown in Algorithm \ref{alg:hpcfe}.
  
 \begin{algorithm}[ht!]
\caption{H-PCFE}\label{alg:hpcfe}

\textbf{Requirements:} Provide Training data set data set, input order of H-PCFE, corresponding parameters and variable bounds\\
compute $g_{0}$ as:\newline
        $g_{0}=\frac{1}{n}\sum_{n}g({z}_{i})$\\
Using the training output data(y), formulate d as\newline
        $\bm{d}={y}-g_{0}$\newline
$\bm{d}\leftarrow [d_{1}\: d_{2}\:... \:d_{n}]^{T}$ \\
Compute the basis function matrix  ${\Psi}$\\
choose a suitable functional form for covariance matrix R\\
The length scale parameter $\theta$ is obtained by maximum likelihood estimate\\
$\mathbf{A}\leftarrow (\mathbf{\Psi}^{T}\mathbf{R}^{-1}\mathbf{\Psi})$, $\mathbf{B}\leftarrow \mathbf{\Psi}^{T}\textbf{R}^{-1}\bm{d}$\\
Obtain $\mathbf{A}'$ and $\mathbf{B'}$ by removing redundants in matrices A and B\\
$\bm{\alpha}_{0} \leftarrow (\mathbf{A'})^{\dagger}\mathbf{B'}$\\
$\mathbf{P}\leftarrow {\mathbf{I}}-(\mathbf{A'})^{-1}\mathbf{A'}$\\
Form the weight matrix $\mathbf{W}_{HA}$ (homotopy algorithm)\\
Obtain $\mathbf{U}$ and $\mathbf{V}$ from the singular value decomposition of $\mathbf{P}\mathbf{W}_{HA}$\\
$\bm{\alpha}_{HA} \leftarrow[\mathbf{V}_{q-r}(\mathbf{U}_{q-r}^{T}\mathbf{V}_{q-r})^{-1}\mathbf{U}_{q-r}^{T}]\bm{\alpha}_{0}$
\end{algorithm}

Once the $\bm{\alpha}_{HA}$ is evaluated, predictive mean and variance at any unknown point $\bm x^*$ is given as: 
\begin{equation}
    \mu({\bm{x}^{*}})=g_{0}+\bm{\Phi}(\bm x^*)\bm{\alpha}_{HA}+\bm{r}(\bm{{x}}^{*})\mathbf{R}^{-1}(\bm{d}-\mathbf{\Psi}\bm{\alpha}_{HA}),
\end{equation}
and
\begin{equation}
     {s}^{2}({\bm{x}^{*}})={\sigma}^{2}\Bigg\{{1}-\bm{r}(\bm{x}^{*})\mathbf{R}^{-1}\bm{r}(\bm{x}^{*})^{T}+\frac{1-{\mathbf{\Psi}}^{T}{\mathbf{R}}^{-1}\bm{r}(\bm{{x}}^{*})^{T}}{{\mathbf{\Psi}}^{T}{\mathbf{R}}^{-1}{\mathbf{\Psi}}}\Bigg\},
\end{equation}
where $\bm \Phi (\bm x^*)$ represents the basis function vector evaluated at $\bm x^*$ and $\bm r (\bm x^*)$ is a vector representing the correlation between $\bm x^*$ and the training inputs.
\par
\noindent
\textbf{Remark 3:} H-PCFE is a hybrid surrogate model that combines PCFE and GP. While PCFE perform a global approximation using basis function, GP tries to capture the local behavior using covariance kernel. Because of this hybridization, H-PCFE yields highly accurate results.
\par
\noindent
\textbf{Remark 4:} The computational cost associated with training H-PCFE can be significant. On one hand, the number of basis functions grows with increase in the dimensionality of the input variables. One the other hand, estimating the length scale parameters $\bm \theta$ involves a numerical optimization problem. As a result, H-PCFE suffers from the \textit{curse of dimensionality} and cannot be directly used for solving high-dimensional reliability analysis problems.

\subsection{Proposed Approach}
Having briefly discussed active subspace, S-PCE and H-PCFE, we now discuss the proposed framework.
The basic premise here is to couple the three methods so that the limitations associated with active subspace, S-PCE and H-PCFE are addressed; this results in a framework that is scalable and can be used for solving high-dimensional reliability analysis problems.

Followings notations discussed before, we consider $\bm X = \left( X_1, X_2, \ldots, X_N \right): \Omega_{\bm X} \in \mathbb R ^N$ to be the inputs and $Y \in \mathbb R$ to be the output variable.
Also, assume that we have a computational model (e.g., finite element model) $f\left(\cdot\right)$ that takes as input realization of variable $\bm X$ and yields the corresponding realization of $Y$
\begin{equation}
    y = f(\bm x),
\end{equation}
where $\bm x$ and $y$, respectively are realization of $\bm X$ and $Y$.
We consider a realistic scenario with (a) $f$ being computationally expensive, (b) gradient of $f$, $\nabla_{\bm x}f$ unavailable and (c) $\bm X$ is high-dimensional. For brevity of representation, we have $f$ instead of $f(\bm x)$
Under such circumstances, the following challenges arise
\begin{itemize}
    \item Computing the low-dimensional manifold by using active subspace becomes intractable from a computational point-of-view. This is because information on the gradient of $f$ is not available and using numerical techniques such as finite difference is expensive as already shown in Eq. \eqref{eq:func_eval_ASS}.
    \item Most of the surrogate models available in the literature suffers from the \textit{curse of dimensionality} and hence, are also intractable when $\bm X$ is high-dimensional. Sparse surrogate models such as S-PCE can be used to learn a mapping between the inputs the output, $\hat {\mathcal M}: \bm X \mapsto Y$; however, results obtained using sparse surrogate models is often not accurate for solving reliability analysis problems.
This is because reliability analysis involves capturing the tail of the response probability distribution.
\end{itemize}
We propose a reliability analysis framework that addresses these challenges and can solve high-dimensional reliability analysis problems.

The first step towards addressing the challenges discussed above is to make active subspace scalable; this will allow the discovery of the low-dimensional manifold.
In this context, we note that, unlike reliability analysis that deals with the tail of the response probability distribution (local measure), $\mathbf C$ (or $\hat {\mathbf C}$ described in Eq. \eqref{montecarlo} ) is a global measure. Therefore, we argue that a sparse learning algorithm like S-PCE is well suited for computing the same. 
In essence, we train a S-PCE model to map between the inputs and the output. Mathematically, 
\begin{equation}
    y = f(\bm x) \approx \hat f(\bm x) = \sum_{\beta \in \mathbb N^N} a_{\beta}\psi_{\beta}(\bm x ).
\end{equation}
We compute the unknown coefficients of PCE by using LAR described in Algorithm \ref{alg:LAR}.
After that, the trained S-PCE model is used for computing $\mathbf C$ and determining the reduced inputs $\bm Z \in \mathbb R^{N_r}$.
We refer to this framework as the sparse active subspace (SAS).
The steps involved in SAS are shown in Algorithm \ref{alg:sas}.
\begin{algorithm}
\caption{Sparse active subspace (SAS)} \label{alg:sas}
\textbf{Choose} the maximum order for PCE.\\
\textbf{Train} the PCE using Algorithm {\ref{alg:LAR}}.\\
\textbf{Select} a suitable value of threshold.\\ \textbf{Project} the inputs onto a low-dimensional manifold using Algorithm {\ref{alg:ASS}} and the trained S-PCE.\\
\end{algorithm}

\begin{figure}
    \centering
    \includegraphics[width = 0.7\textwidth]{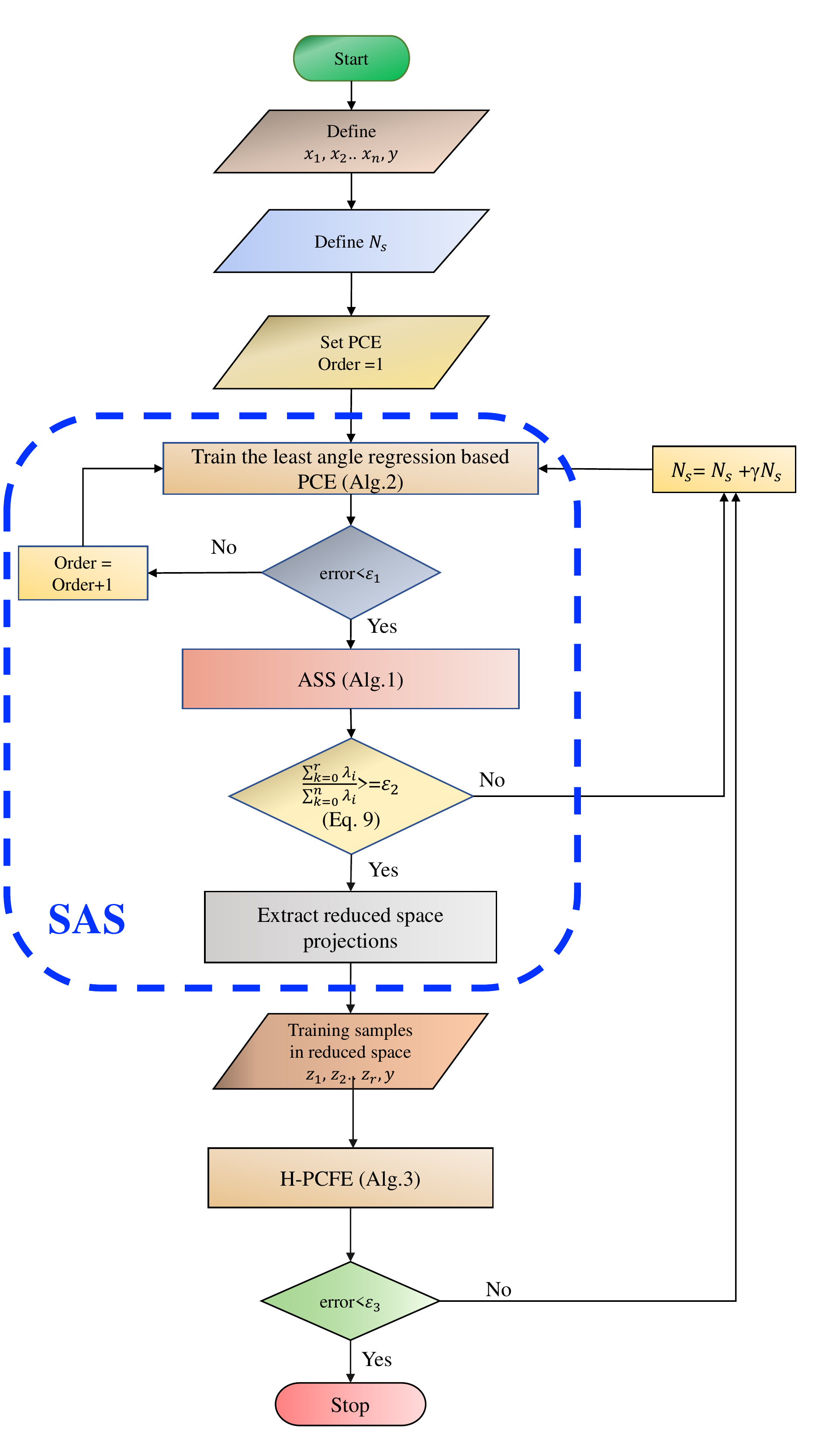}
    \caption{Flow chart describing step-wise procedure for the proposed approach. The portion marked in blue indicates sparse active subspace. We refer to the overall framework as SAS-HPCFE.}
    \label{fig:Flowchart}
\end{figure}

Once the reduced inputs $\bm Z$ have been identified, the remainder of the task is straightforward; we train a surrogate model between the reduced inputs and the output, $\mathcal M_1: \bm Z \mapsto Y$. Since the objective of this paper is to develop a framework capable of solving reliability analysis problems, we use H-PCFE as a surrogate model.
As already stated, H-PCFE performs a bi-level approximation and hence, yields highly accurate results. Moreover, with the low-dimensional $\bm Z$ being the inputs to the H-PCFE, the issue with \textit{curse of dimensionality} does not arise. We refer to the overall framework as SAS-based H-PCFE (SAS-HPCFE). 
For clarity of readers, a flowchart depicting the overall framework is shown in \autoref{fig:Flowchart}.

\section{Numerical examples}\label{sec:ne}
This section presents three numerical examples to illustrate the application and utilization of the proposed SAS-HPCFE.
The problems selected involve both analytical and numerical limit-state functions.
Although the proposed approach is equally applicable with all design of experiments schemes \cite{bhattacharyya2018critical}, quasi-random sample points have been used in this study. Among different quasi-random sequences available in the literature \cite{sobol1976uniformly,bratley1988algorithm,galanti1997low}, Sobol sequence has been used because of its superior convergence rate \cite{sobol1976uniformly}.
The objective in all the problems is to compute the probability of failure ($P_f$) and the reliability index ($\beta$),
\begin{equation}
    \beta = \Phi^{-1}\left( 1 - P_f \right),
\end{equation}
where $\Phi^{-1}$ indicates inverse normal cumulative distribution function (CDF).
For illustrating the accuracy of the proposed approach, benchmark results using crude MCS have been generated.
Efficiency, on the other hand, is compared in terms of the number of actual function evaluations.
For illustrating the superiority of the proposed approach, we compare the results obtained using the proposed approach with those obtained using First-Order Reliability Method (FORM), Second-Order Reliability Method (SORM) and S-PCE. All the methods have been implemented using MATLAB \cite{higham2016matlab}.
For FORM and SORM, already available FERUM \cite{bourinet2009review} package has been used.

\subsection[Example 1: Sobol function]{Example 1:Sobol function \cite{marrel2008efficient}}
As the first example, we consider an analytical example where the limit-state function is defined in terms of the Sobol function \cite{marrel2008efficient}
\begin{equation}\label{sob}
   Y=g(\bm{X})=\prod_{i=1}^{m}\frac{|4X_{i}+2|+a_{i}}{1+a_{i}}-b,
\end{equation}
where $\bm X=(X_{1},X_{2},...,X_{m})^{T}$ is a set of m-dimensional independent random variables distributed uniformly in [0,1]. 
The problem has two distinct features making it an ideal choice for illustrating the proposed SAS-HPCFE.
First, the constants $a=[a_{1},a{2},...,a_{m}]$ are non negative parameters which denote the influence of an input variable $X_{i}$ on output $Y$. A lower value of  $a_{i}$ indicates higher influence of variable $X_{i}$ on the output.
In other words, the dimensionality of the active subspace is known a-priori and hence, it is possible to examine whether the proposed SAS is able to capture the correct subspace or not.
Secondly, the functional form shown in Eq. \eqref{sob} is highly nonlinear, and hence, it is challenging for a surrogate model to approximate the same. Note that popularly used model reduction techniques such as principal component analysis will not be able to reduce the dimensionality as the input variables are independent.
The constant $b$ in Eq. \eqref{sob} prescribes the threshold value, which determines the magnitude of failure probability. Here we analyze the performance of the proposed approach for three different dimensions $(m=10,40,100)$. In all three cases, the set of $a_{i}$ values are chosen as $a={(1,1,\underbrace{500,....,500}_{m-2} )}^T$ and threshold $b=0.35$. These values indicate that the true subspace can be approximated to two.

For solving these three cases using the proposed SAS-HPCFE, we first build the SAS by using S-PCE and the active subspace algorithm discussed before.
For S-PCE, the maximum polynomial order is set to be $5$, and the number of training samples is decided based on convergence study. Once the S-PCE is trained, we utilize the active subspace algorithm with a threshold of 0.98 for computing the optimal subspace and projecting the high-dimensional input $\bm X$ on the low-dimensional manifold identified by the active subspace; this yields the reduced-order input $\bm Z$. 
Finally, we train the high-fidelity H-PCFE model to map between the low-dimensional input $\bm Z$ and the output $Y$.
We utilize the same training samples used in S-PCE by projecting it onto the low-dimensional manifold; this ensures that no additional actual function evaluations are needed.

\begin{figure}[ht!]
    \centering
    \subfigure[]{
    \includegraphics[width = 0.48\textwidth]{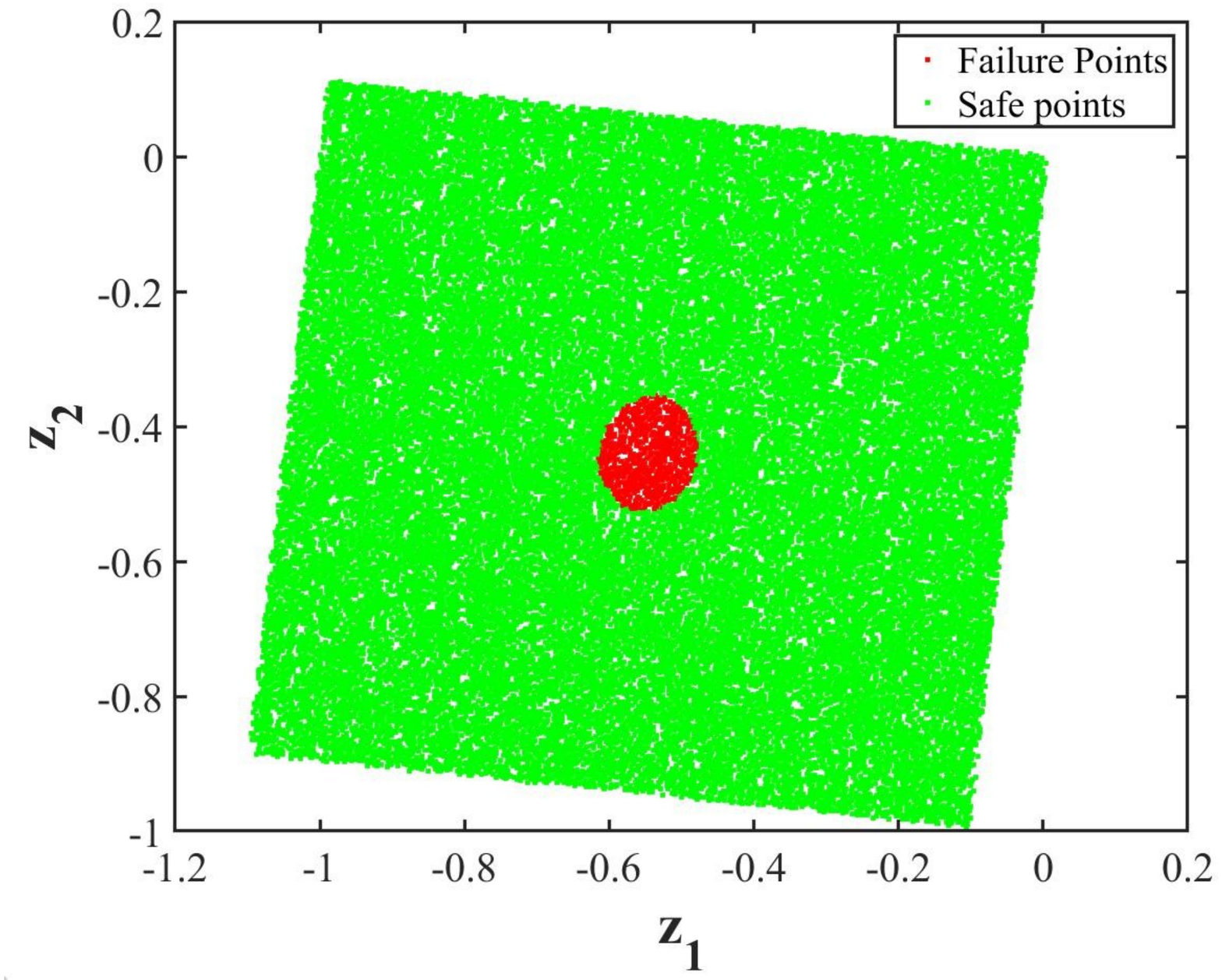}}
    \subfigure[]{  
    \includegraphics[width = 0.48\textwidth]{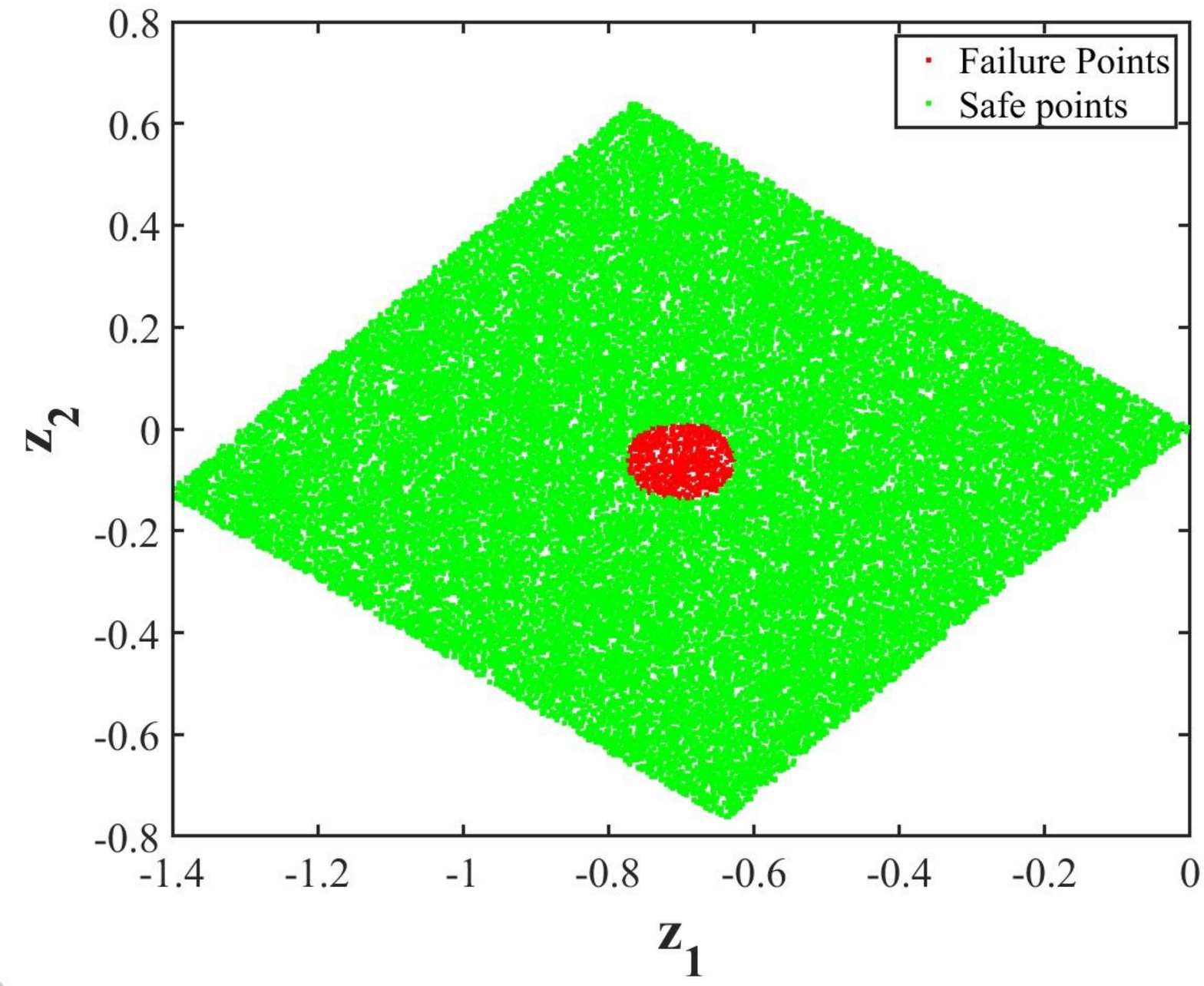}}
    \subfigure[]{  
    \includegraphics[width = 0.48\textwidth]{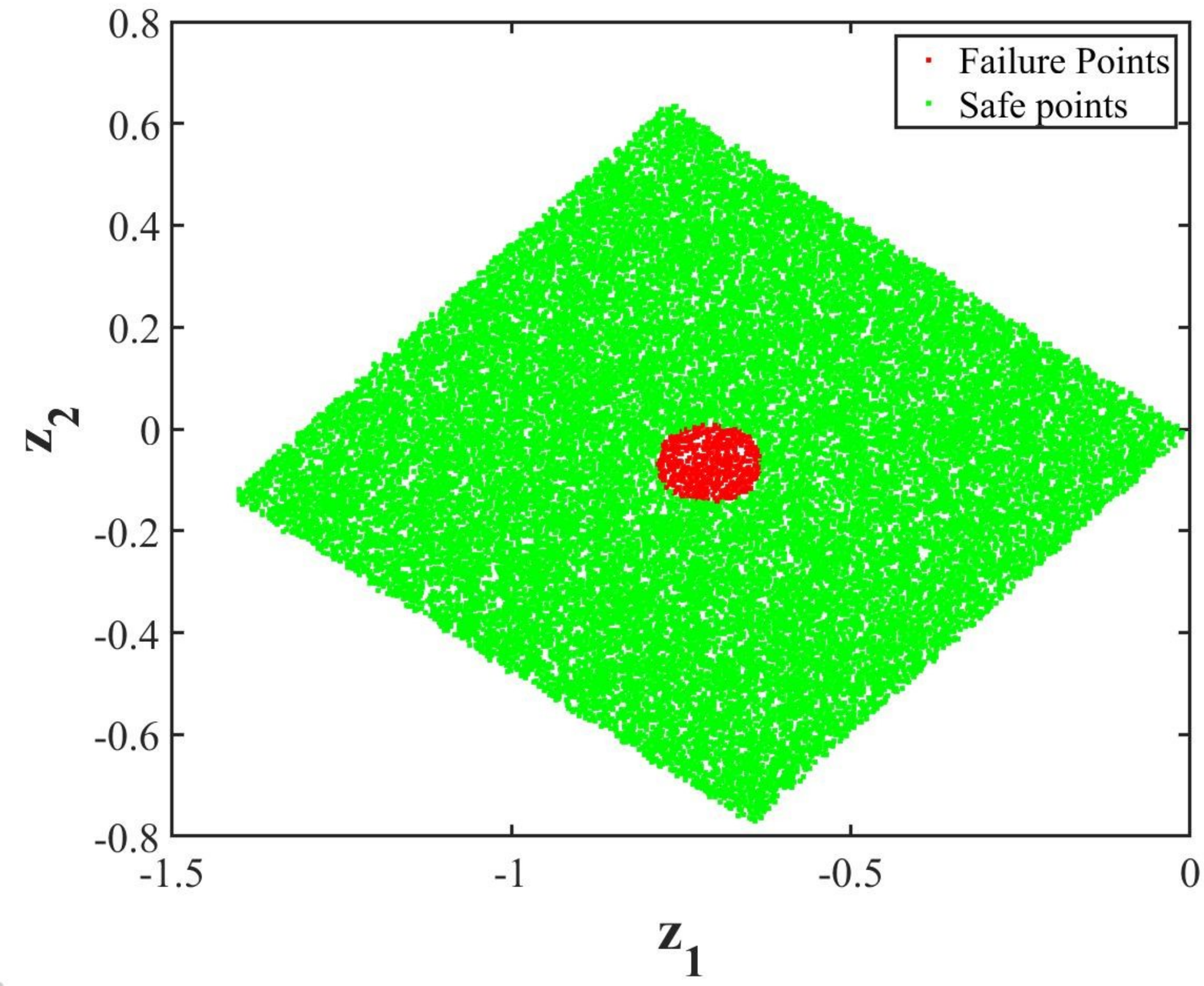}}
    \caption{Classification of safe and failure points predicted from H-PCFE estimation  within the PCE-based subspace for the three cases of sobol function (a) $m=10$,(a) $m=40$, and (c) $m=100$}
    \label{fig:102}
\end{figure}

Before proceeding with the reliability analysis results, we first examine the active subspace identified by the proposed SAS.
\autoref{fig:102} shows the classification of estimated safe points and failure points within the reduced subspace. Here $\bm{Z_{1}}$ and  $\bm{Z_{2}}$ are the transformed coordinates of reduced subspace. We observe that for all three cases, the proposed SAS has identified the active subspace dimension to be two, which exactly matches the actual subspace dimensionality.
This indicates the effectiveness of the proposed approach in accurately identifying the active subspace.

Having illustrated the effectiveness of the proposed SAS in identifying the active subspaces, we next examine the performance of SAS-HPCFE in solving reliability analysis problems. 
First, the relatively easier case with $m=10$ is considered. \autoref{table1} shows the results obtained using MCS, SAS-HPCFE, FORM, SORM and S-PCE. Compared to MCS results $\left( P_f = 0.0177, \beta = 2.1038 \right)$, S-PCE $\left( P_f = 0.01839, \beta = 2.0882 \right)$ and the proposed SAS-HPCE $\left( P_f = 0.01846, \beta = 2.0866 \right)$ yield the most accurate results with an prediction error of $0.74\%$ and $0.82\%$, respectively.
However, the number of training samples needed for S-PCE is almost 60\% more as compared to the proposed SAS-HPCFE.
This is because S-PCE is not efficient in capturing local behavior (e.g., reliability) and needs more training samples.
FORM $\left( P_f = 0.5680, \beta = -0.1713 \right)$ and SORM $\left( P_f = 0.6775, \beta = -0.4607 \right)$ yield prediction error of 108.14\% and 121.89\%, respectively.
The erroneous results obtained using FORM and SORM are due to the high nonlinearity of the limit-state function.

\begin{table}[ht!]
    \centering
    \caption{Probability of failure for Sobol with 10 variables obtained by various methods. $\beta_e$ indicates the reliability index obtained using MCS.}
    \label{table1}
\begin{tabular}{lcccc} 
\hline
\textbf{Method }& \textbf{Reliability Index} & \textbf{Failure Provability} & \textbf{$N_{s}$}
&$\epsilon=\frac{|\beta_{e}-\beta|}{\beta_{e}}\times100$\\\hline
MCS           &  2.1038  &   0.0177  &  $10^{5}$ & -\\ \hline\hline
FORM          &  -0.1713 &   0.5680  &   45 & 108.1424 \\ 
SORM          & -0.46065  &   0.67747 &  65 & 121.8961\\ 
S-PCE           &  2.0882  &   0.01839 &  1300 & 0.74\\ 
SAS-HPCFE    &  2.0866  &   0.01846 &  800 & 0.82\\ 

\hline 
\end{tabular}
\end{table}

Next, we consider a more challenging scenario with $m=40$. 
\autoref{table2} shows the results obtained using different methods.
We observe that the performance of S-PCE $\left( P_f = 0.0573, \beta = 2.5284 \right)$ deviates with increase in the dimensionality of the input variables. This is due to the fact that S-PCE fails to capture the tail region of the response probability distribution. Performance of S-PCE with an increase in the number of training samples was also examined; however, no significant improvement beyond the reported results was observed. For brevity of representation, the case studies are not shown here. The proposed SAS-HPCE $\left( P_f = 0.1809, \beta = 2.0949 \right)$, on the other hand, yields equally accurate results with prediction error of 0.31\%. 
FORM $\left( P_f = 0.5806, \beta = -0.2036 \right)$ and SORM $\left( P_f = 0.01271, \beta = 1.1402 \right)$ yields prediction error of 109.69\%
and 45.74\%, respectively.
Similar to the previous case, inaccuracies in FORM and SORM predictions is due to the high nonlinearity in the limit-state function.
Also, note that with an increase in the dimensionality of the input variables from 10 to 40, the number of actual function evaluations for FORM and SORM increases by approximately 4 times and 13 times, respectively. The proposed SAS-HPCFE, on the other hand, requires an almost equal number of training samples for both cases. This illustrates the scalability of the proposed approach.

\begin{table}[ht!]
    \centering
    \caption{Probability of failure for Sobol with 40 variables obtained by various methods.  $\beta_e$ indicates the reliability index obtained using MCS.}
    \label{table2}
\begin{tabular}{lcccc} 
\hline
\textbf{Method }& \textbf{Reliability Index} & \textbf{Failure Provability} & \textbf{$N_{s}$} &$\epsilon=\frac{|\beta_{e}-\beta|}{\beta_{e}}\times100$\\
\hline
MCS           &  2.1015  &   0.0178 &   $10^{5}$ & - \\ \hline\hline
FORM          &  -0.2036 &   0.5806  &  182  & 109.6883\\ 
SORM          &  1.1402  &   0.01271 &  860  & 45.7435 \\ 
S-PCE           &  2.5284  &   0.00573 &  1300 &  20.3141\\ 
SAS-HPCFE    &  2.0949  &   0.01809 &  900  & 0.3141\\ 
\hline 
\end{tabular}
\end{table}

Lastly, we consider the case with $m=100$. Results obtained using different methods are shown in \autoref{table3}. Even for this case, SAS-HPCFE $\left( P_f = 0.0189, \beta = 2.0949 \right)$ yields accurate result with prediction error of 1\%.
FORM $\left( P_f = 0.6062, \beta = -0.2694 \right)$, SORM $\left( P_f = 0.0051, \beta = 2.5692 \right)$ and S-PCE $\left( P_f = 0.002, \beta = 2.8782 \right)$ yield inaccurate results with prediction error of 112.85\%, 22.52\% and 37.26\%, respectively. Perhaps more importantly, we observe that the number of actual function evaluations for FORM and SORM has increased to 422 and 5150 respectively. The proposed approach, one the other hand, yields accurate results with minimal increase in the number of training samples. This establishes the scalability of the proposed approach.

\begin{table}[ht!]
    \centering
    \caption{Probability of failure for Sobol with 100 variables obtained by various methods.  $\beta_e$ indicates the reliability index obtained using MCS.}
    \label{table3}
\begin{tabular}{lcccc} 
\hline
\textbf{Method }& \textbf{Reliability Index} & \textbf{Failure Provability} & \textbf{$N_{s}$}&$\epsilon=\frac{|\beta_{e}-\beta|}{\beta_{e}}\times100$\\
\hline
MCS           &  2.0969  &   0.0180 &   $10^{5}$ &-\\ \hline\hline
FORM          &  -0.2694 &   0.6062  &  422 & 112.8475\\ 
SORM          &  2.5692  &   0.0051  &  5150 & 22.5237\\ 
S-PCE           &  2.8782  &   0.002   &  1300 & 37.2598\\ 
SAS-HPCFE    &  2.0757  &   0.0189 &  1100 & 1.0110\\ 
\hline 
\end{tabular}
\end{table}

\subsection[Example 2: A composite beam]{Example 2:A composite beam \cite{xiao2014novel}}
\begin{figure}[ht!]
    \centering
    \includegraphics[width=.8\textwidth]{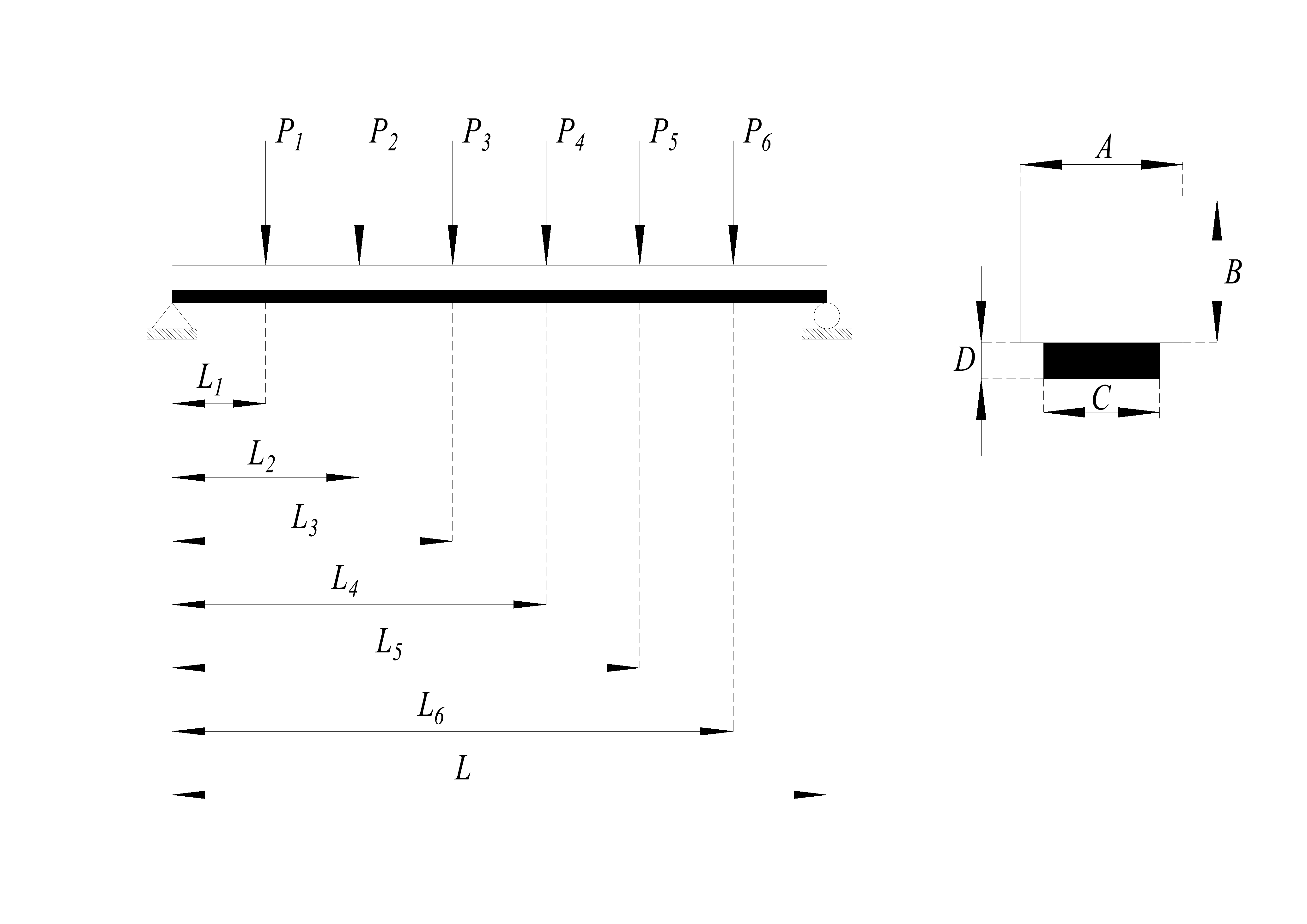}
    \caption{ Composite beam considered in Example 2.}
    \label{fig:beam}
\end{figure}

As the second example, we consider a composite beam shown in Fig. \ref{fig:beam}.
The beam has a cross-sectional area $A(mm)\times B(mm)$. An aluminium plate of cross-section $C(mm)\times D(mm)$ is securely fastened to the bottom face of the beam. Young's modulus of the beam material and aluminum are given by $E_{w}$ and $E_{a}$, respectively. The beam is spanned by a length, L and six different external loads, $P_{1}$, $P_{2}$, $P_{3}$, $P_{4}$, $P_{5}$, and $P_{6}$, are applied at distances, $L_{1}$, $L_{2}$, $L_{3}$, $L_{4}$, $L_{5}$, and $L_{6}$ measured from left end. The maximum allowable stress is given by S. Limit state function of the composite beam is defined in terms of aforementioned twenty independent variables  
The limit state function is represented as
\begin{equation}
    g(x)=S-\sigma
\end{equation}
Where maximum developed stress in the beam, $\sigma$ is described as
\begin{equation}\label{f2}
   \sigma = \frac{\Bigg[\frac{\sum_{i=1}^{6}{P_{i}(L-L_{i})}}{L}{L_{3}}-P_{1}(L_{3}-L_{1})-P_{2}(L_{3}-L{2})\Bigg]{K}}{\frac{1}{12}{A}{B}^{3}+AB\{K-0.5{B}\}^{2}+\frac{1}{12}\frac{E_{a}}{E_{w}}{C}{D}^{3}+\frac{E_{a}}{E_{w}}{D}{C}\{B+0.5D-K\}^{2}},
\end{equation}
where
\begin{equation}
    K=\Bigg[\frac{0.5{A}{B}^{2}+\frac{E_{a}}{E_{w}}{D}{C}(B+0.5D)}{{A}{B}+\frac{E_{a}}{E_{w}}{D}{C}}\Bigg].
\end{equation}

Detailed information of theses random variables are provided in \autoref{table4}.

\begin{table}[ht!]
    \centering
    \caption{Distribution type and distribution parameters for composite beam in Example 2.
}
    \label{table4}
\begin{tabular}{lccccc} 
\hline
{Variable no.} & {Variable} & {Mean} & {SD} & {Distribution} & {Supporting Interval}\\
\hline
1& A (mm) & 100 & 0.2 & Normal & [99.4, 100.6]\\
2& B (mm) & 200 & 0.2 & Normal & [199.4, 200.6]\\
3& C (mm) & 80 & 0.2 & Normal & [79.4, 80.6]\\
4& D (mm) & 20 & 0.2 & Normal & [19.4, 20.6]\\
5& $L_{1}$ (mm) & 200 & 1 & Normal & [197, 203]\\
6& $L_{2}$ (mm) & 400 & 1 & Normal & [397, 403]\\
7& $L_{3}$ (mm) & 600 & 1 & Normal & [597, 603]\\
8& $L_{4}$ (mm) & 800 & 1 & Normal & [797, 803]\\
9& $L_{5}$ (mm) & 1000 & 1 & Normal & [997, 1003]\\
10& $L_{6}$ (mm) & 1200 & 1 & Normal & [1197, 1203]\\
11& L (mm) & 1400 & 2 & Normal & [1394, 1406]\\
12& $P_{1}$ (kN)& 15 & 1.5 & Gumbel & [5, 19]\\
13& $P_{2}$ (kN)& 15 & 1.5 & Gumbel & [5, 19]\\
14& $P_{3}$ (kN)& 15 & 1.5 & Gumbel & [5, 19]\\
15& $P_{4}$ (kN)& 15 & 1.5 & Gumbel & [5, 19]\\
16& $P_{5}$ (kN)& 15 & 1.5 & Gumbel & [5, 19]\\
17& $P_{6}$ (kN)& 15 & 1.5 & Gumbel & [5, 19]\\
18& Ea (GPa)& 70 &  7 & Normal & [49, 91]\\
19& Ew (GPa)& 8.75 & 0.875 & Normal & [6.125, 11.375]\\
20& S (MPa)& 21 & 2.1 & Gumbel & [16, 35]\\
\hline 
\end{tabular}
\end{table}

To solve this problem using the proposed SAS-HPCFE, we first employ SAS to identify the low-dimensional manifold; this involves training a S-PCE and then using it within the active subspace algorithm.
For training S-PCE, we set the maximum order of the basis function to five. The required number of training samples were computed based on convergence study. Once we have the low-dimensional manifold, we project the high-dimensional input onto the low-dimensional manifold by using the SAS. Finally, we train H-PCFE to map between the low-dimensional input $\bm Z$ and the output $Y$. Similar to the previous example, we project the training points used in S-PCE on to the low-dimensional manifold and reuse it for training the H-PCFE. Consequently, no additional actual function evaluations are needed for training H-PCFE.
Unlike the previous example, the original dimension of the subspace is not known for this problems. Therefore, it is not possible to comment on the same. Nevertheless, using a threshold of 0.98, SAS reduces the input dimensionality from 20 to 3. Note that using conventional dimensionality reduction methods such as PCA, it will not be possible to reduce the dimension as the variables are independent.

\autoref{table5} shows the reliability analysis results obtained using different methods.
Probability of failure, reliability index, number of actual function evaluations and prediction error are reported.
Crude MCS with $10^6$ simulations is considered as the benchmark solution. The higher number of MCS samples used for this problem is because of the low probability of failure. We observe that the proposed SAS-HPCFE $\left( P_f = 0.00167, \beta = 2.9346 \right)$ yields highly accurate results with predictive error of 0.8\%.
FORM $\left( P_f = 0.00087, \beta = 3.1427 \right)$ and S-PCE $\left( P_f = 0.00057, \beta = 3.2535 \right)$ yield predictive accuracy of 7.95\% and 11.76\% respectively.
As for computational cost, the proposed SAS-HPCFE $(N_s = 800)$ is found to be most efficient followed by S-PCE $(N_s = 1000)$ and FORM $(N_s = 3194)$.
SORM didn't converge even after $10^4$ function evaluations.

\begin{table}[ht!]
    \centering
    \caption{Probability of failure for Composite beam obtained by various methods.  $\beta_e$ indicates the reliability index obtained using MCS.}
    \label{table5}
\begin{tabular}{lcccc} 
\hline
\textbf{Method }& \textbf{Reliability Index} & \textbf{Failure Provability} & \textbf{$N_{s}$}&$\epsilon=\frac{|\beta_{e}-\beta|}{\beta_{e}}\times100$\\
\hline
MCS           &  2.9112  &   0.0018 &   $10^{6}$ & -\\ \hline\hline
FORM          &  3.1427 &   0.00087  &  3194 & 7.9520\\ 
SORM          &  -  &  -  &  - & -\\ 
S-PCE           &  3.2535  &  0.00057   &  1000 & 11.7580 \\ 
SAS-HPCFE    &  2.9346  &   0.00167 &  800 &  0.8038 \\ 
\hline 
\end{tabular}
\end{table}

\subsection[Twenty five element space truss]{Twenty five element space truss \cite{patelli2010cossan}}
\begin{figure}
    \centering
    \subfigure[]{
    \includegraphics[width=.9\textwidth]{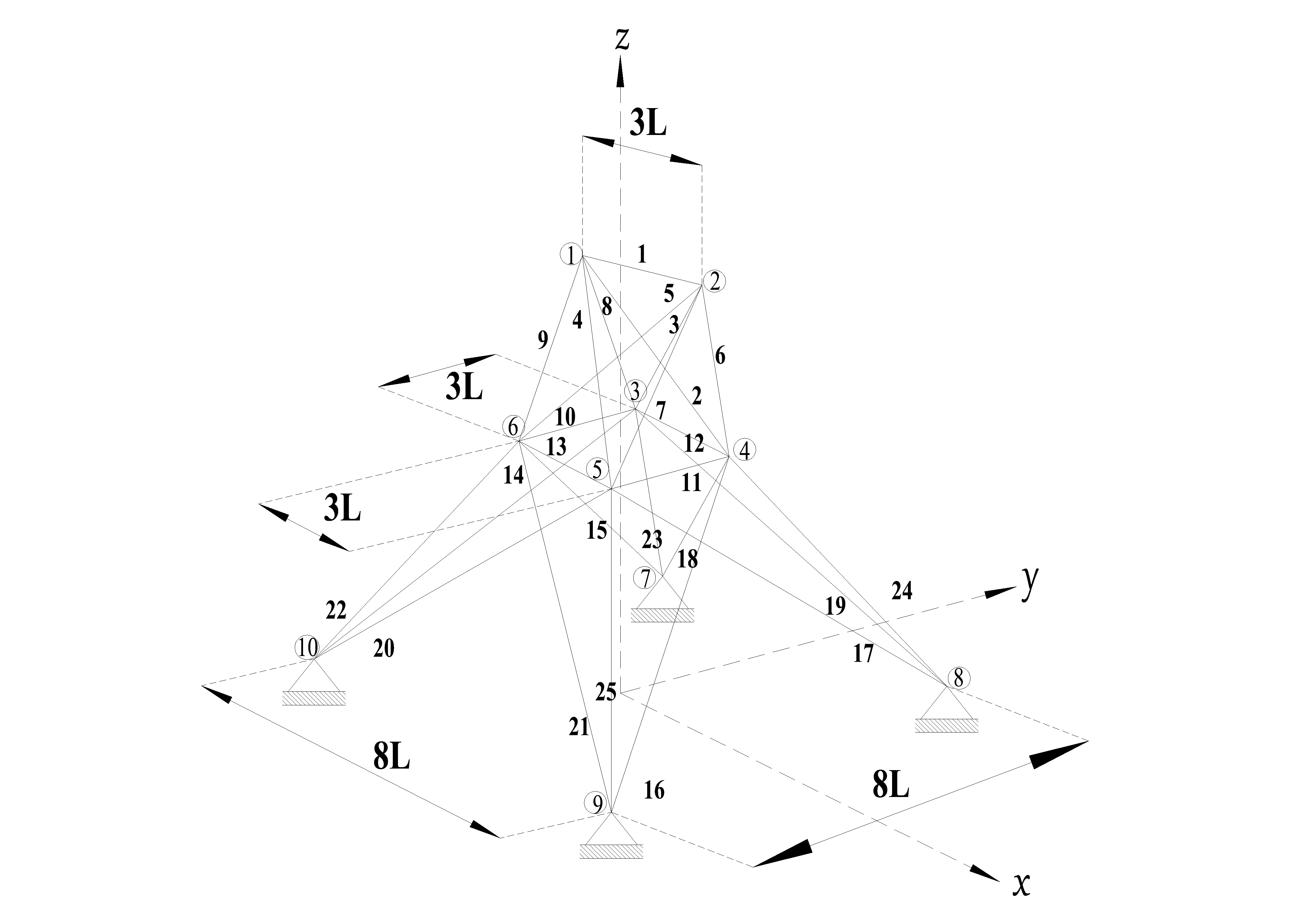}}
    \subfigure[]{
    \includegraphics[width=.6\textwidth]{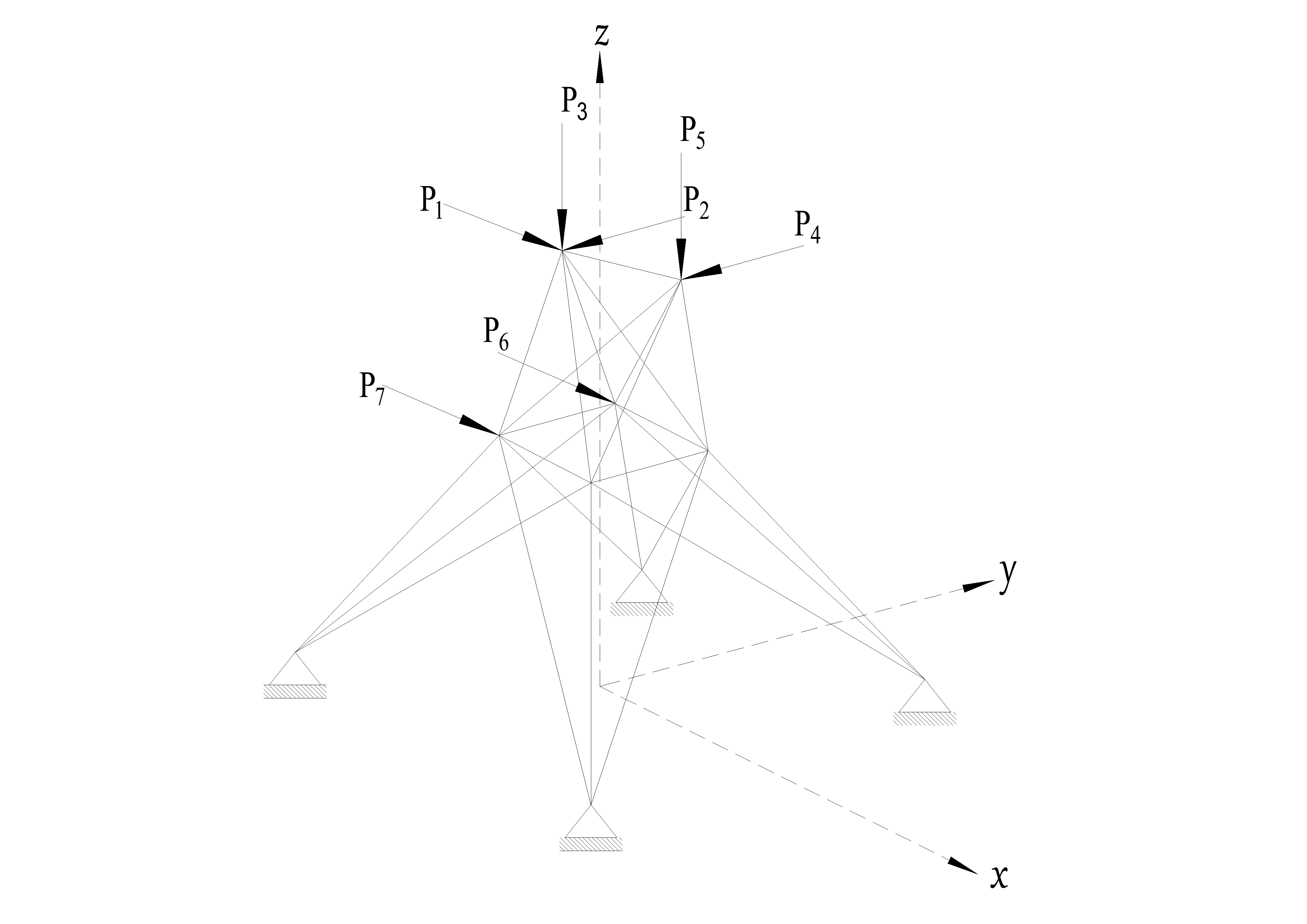}}
    \caption{Truss with twenty five elements considered in Example 3; (a) dimensional details along with node and element numbers, (b) loading details}
    \label{fig:truss2}
\end{figure}

As the last example, we consider a practical example involving a space truss.
The truss considered has $25$ elements and $10$ nodes.
A schematic representation of the space truss is shown in Fig. \ref{fig:truss2}(a).
The system is subjected to 5 horizontal ($P_1, P_2, P_4, P_6$ and $P_7$) and 2 vertical loads ($P_3$ and $P_5$). A schematic representation of the the space truss with location of the external loads is shown in Fig \ref{fig:truss2}(b).

We have considered Young's modulus, the cross-section of the members and external forces to be stochastic. Altogether 33 random variable describes the truss structure. Detailed information on these random variables is presented in the \autoref{table6}.

\begin{table}[ht!]
    \centering
    \caption{Distribution type and distribution parameters for truss in Example 3.
}
    \label{table6}
\begin{tabular}{lcccc} 
\hline
{Variable no.} & {Variable} & {Mean} & {SD} & {Distribution} \\
\hline
1& $P_{1}$ (N) & 1000 & 100 & Lognormal \\
2-5& $P_{2}-P_{5}$ (N) & 10,000 & 500 & Normal \\
6& $P_{6}$ (N) & 600 & 60 & Lognormal \\
7& $P_{7}$ (N) & 500 & 50 & Lognormal\\
8& $E $ $\mathrm{(N/m^{2})}$ & $10^{7}$ & $5\times10^{5}$ & Lognormal\\
9& $A_{1} $ $\mathrm{(m^{2})}$ & 0.4 & 0.04 & Lognormal \\
10-13& $A_{2}-A_{5}$ $ \mathrm{(m^{2})}$ & 0.1 & 0.01 & Lognormal\\
14-17& $A_{6}-A_{9}$ $ \mathrm{m^{2})}$  & 3.4 & 0.34 & Lognormal\\
18-19& $A_{10}-A_{11}$ $ \mathrm{(m^{2})}$  & 0.4 & 0.04 & Lognormal\\
10-21& $A_{12}-A_{13}$ $ \mathrm{(m^{2})}$   & 1.3 & 0.13 & Lognormal \\
22-25& $A_{14}-A_{17}$ $ \mathrm{(m^{2})}$  & 0.9 & 0.09 & Lognormal \\
26-29& $A_{18}-A_{21}$ $ \mathrm{(m^{2})}$  & 1 & 0.1 & Lognormal \\
30-33& $A_{21}-A_{25}$ $\mathrm{(m^{2})}$  & 3.4 & 0.34 & Lognormal \\

\hline 
\end{tabular}
\end{table}

Another essential aspect when it comes to reliability analysis is defining the limit-state function. We have defined the limit-state function in terms of the max deformation the system is undergoing.
Mathematically, this is defined as
\begin{equation}\label{Eq:lmfunctiontruss}
    g(\bm{X})=u_{0}-max(u_{1},u_{2}),
\end{equation}
where $u_{1}$ and $u_{2}$ in the Eq. \eqref{Eq:lmfunctiontruss} represent the peak displacements in horizontal and vertical directions and $u_{0}$ denotes the maximum allowable displacement. For this example, we have considered $u_0 = 0.4$.

For solving this problem using the proposed SAS-HPCFE, we follow the same procedure as examples 1 and 2 with the same setup. Using SAS, we obtained the dimensionality of the reduced space to be seven.
However, for this problem also, the exact dimensionality of the active subspace is unknown, and hence, the identified active subspace could not be verified.

\begin{table}[ht!]
    \centering
    \caption{Probability of failure for twenty five element truss obtained by various methods. $\beta_e$ represents the reliability obtained using MCS.}
    \label{table7}
\begin{tabular}{lcccc} 
\hline
\textbf{Method }& \textbf{Reliability Index} & \textbf{Failure Provability} & \textbf{$N_{s}$} &$\epsilon=\frac{|\beta_{e}-\beta|}{\beta_{e}}\times100$\\
\hline
MCS           &  1.2443  &   0.1067 &   $10^{5}$ & -\\ \hline\hline
FORM          &  1.5816 &   0.005687  &  194 & 27.1076\\ 
SORM          &  1.5521  &   0.006032  &  594 & 24.7368\\ 
PCE           &  1.2925 &   0.0891   &  1100 & 3.8737\\ 
SAS-H-PCFE    &  1.2527  &   0.10515 &  1000 & 0.6751\\ 
\hline 
\end{tabular}
\end{table}

\autoref{table7} shows the reliability analysis results obtained using MCS, FORM, SORM, S-PCE and the proposed SAS-HPCFE. 
Compared to benchmark solutions obtained using MCS with $10^5$ samples $\left( P_f = 0.1067, \beta = 1.2443 \right)$, the proposed approach $\left( P_f = 0.1052, \beta = 1.2527 \right)$ yields the best result with prediction error or 0.67\%. Results obtained using S-PCE $\left( P_f = 0.0891, \beta = 1.2925 \right)$ are also reasonable; although it require 100 additional training samples. FORM $\left( P_f = 0.005687, \beta = 1.5816 \right)$ and SORM $\left( P_f = 0.006032, \beta = 1.5521 \right)$ yield prediction error of 27.1\% and 24.7\%, respectively.

\section{Conclusions}\label{sec:conclusions}
In this paper, we have presented a novel approach to evaluate the reliability of high-dimensional systems by using active subspace algorithm and surrogate models. We argue that active subspace algorithm and surrogate models on its own are often not suitable for solving real-life problems with high-dimensional inputs; this is because both active subspace and surrogate models suffer from the \textit{curse of dimensionality}. To address this issue, we propose a novel framework where active subspace and surrogate models assist each other. To be specific, we propose to combine sparse polynomial chaos expansion (S-PCE) with active subspace algorithm. We note that S-PCE being a global surrogate model is often not accurate enough for solving reliability analysis problems as it involves capturing the tail region of the response probability distribution; however, S-PCE is good at capturing global behavior (e.g., capturing covariance). We exploit this property to first build a S-PCE and then use it for determining the low-dimensional manifold by employing active subspace algorithm. The resulting algorithm is referred to as the sparse active subspace (SAS). Thereafter a highly accurate surrogate referred to as the H-PCFE is used for mapping the reduced input variables residing in the low-dimensional manifold (identified using SAS) with the output response. The overall framework is referred to as SAS based H-PCFE (SAS-HPCFE)

The proposed SAS-HPCFE is used for solving three numerical examples involving analytical and numerical limit-state functions. Results obtained have been compared with different state-of-the-art reliability analysis methods. For all the examples, the proposed approach yields highly accurate results outperforming popular reliability analysis methods. As for efficiency, we found the proposed approach to efficient (indicated by the number of function evaluations) and scalable (increase in number of training samples is minimal with increase in number of random variables).

Despite the success of the proposed approach, there are a few limitations that needs to be addressed.
\begin{itemize}
    \item Within the proposed SAS-HPCFE, we separately train PCE, identifies active subspace and train H-PCFE. The resulting framework can be accelerated if the three steps can be carried out simultaneously.
    \item Active subspace is a linear manifold learning algorithm. Employing a nonlinear manifold learning algorithm has the potential to further reduce the dimensionality. Obviously, a nonlinear manifold learning algorithm has its own challenges. But given the huge potential, this is an area that needs to be further investigated.
    \item All the three problems solved in this paper involve static problems. How the proposed approach will perform for dynamical systems needs further investigation.
\end{itemize}
In future, research will be conducted to explore some of this research directions.

\section*{Acknowledgements}
NN acknowledges the support received for Ministry of Education in form of Ph.D. scholarship. SC acknowledges the financial support received from IIT Delhi and I-Hub Foundation for Cobotics (IHFC) in form of seed grant.


\end{document}